\def\BLIND{0} 

\documentclass[a4paper,twoside]{article}
\usepackage{bm}
\usepackage{booktabs}
\usepackage{multirow}
\usepackage{natbib}
\usepackage{fix-cm}

\usepackage{epsfig}
\usepackage{subcaption}
\usepackage{calc}
\usepackage{amssymb}
\usepackage{amstext}
\usepackage{amsmath}
\usepackage{amsthm}
\usepackage{multicol}
\usepackage{pslatex}
\usepackage{apalike}
\usepackage{algorithm2e}
\usepackage[bottom]{footmisc}

\usepackage{balance}

\usepackage{SCITEPRESS}     

\begin{document}

\title{Dynamic Graph Representation with Contrastive Learning for Financial Market Prediction: Integrating Temporal Evolution and Static Relations}

\ifnum\BLIND=1
    \author{Anonymous submission 
    \\ } 
\else
    \author{\authorname{Yunhua Pei\sup{1}\orcidAuthor{0000-0003-2906-0827}, Jin Zheng\sup{2}\orcidAuthor{0000-0002-1783-1375} and John Cartlidge\sup{2}\orcidAuthor{0000-0002-3143-6355}}
    \affiliation{\sup{1}School of Computer Science, University of Bristol, UK}
    \affiliation{\sup{2}School of Engineering Mathematics and Technology, University of Bristol, UK}
    \email{\{ge22472, jin.zheng, john.cartlidge\}@bristol.ac.uk}}
\fi

\keywords{Contrastive learning, Financial market forecasting, Graph neural networks, Temporal graph learning.}

\abstract{Temporal Graph Learning (TGL) is crucial for capturing the evolving nature of stock markets. Traditional methods often ignore the interplay between dynamic temporal changes and static relational structures between stocks. To address this issue, we propose the Dynamic Graph Representation with Contrastive Learning (DGRCL) framework, which integrates dynamic and static graph relations to improve the accuracy of stock trend prediction. Our framework introduces two key components: the Embedding Enhancement (EE) module and the Contrastive Constrained Training (CCT) module. The EE module focuses on dynamically capturing the temporal evolution of stock data, while the CCT module enforces static constraints based on stock relations, refined within contrastive learning. This dual-relation approach allows for a more comprehensive understanding of stock market dynamics. Our experiments on two major U.S. stock market datasets, NASDAQ and NYSE, demonstrate that DGRCL significantly outperforms state-of-the-art TGL baselines. Ablation studies indicate the importance of both modules.  Overall, DGRCL not only enhances prediction ability but also provides a robust framework for integrating temporal and relational data in dynamic graphs. Code and data are available for public access.}

\onecolumn \maketitle \normalsize \setcounter{footnote}{0} \vfill

\section{\uppercase{Introduction}}
\label{sec:introduction}


The goal of predicting stock movements attracts much attention, as success offers the opportunity to generate substantial investment returns. Previous research can generally be divided according to the input features used in the model. Historical price is the most common and basic input feature, including the open price, close price, highest price, lowest price, and trading volume (OHLCV). Many predictions have been made based solely on historical prices, including applying empirical mode decomposition with factorization machine-based neural networks \citep{zhou2019emd2fnn} and forming a stochastic recurrent network with seq2seq architecture and attention mechanism \citep{ijcai2019p514}. More attempts include adding technical indicators such as achieving Long Short-Term Memory (LSTM) with the attention mechanism \citep{chen2019exploring}, creating multitask RNNs with high-order Markov random fields \citep{li2019multi}. 

Graphs, which represent information through entities, attributes, and their relationships, have recently gained popularity. This is due to their effectiveness in illustrating the connections between stocks and their various attributes. By casting a graph, relevant information transfer between stocks and stock attributes from other channels can be studied. For instance, the attention mechanism is used to build a market knowledge graph to contain dual-type entities and mixed relationships \citep{zhao2022stock}, GCNs with temporal graph convolution are performed on data in a rolling window \citep{matsunaga2019exploring}, and incorporating emotional factors within a non-stationary Markov chain model \citep{liu2022forecasting}. These graph-based models leverage the structure of GNNs to iteratively aggregate, showing potentially powerful performance compared with non-graph-based models.

Although previous research has delved into comprehensive investigations of the stock market, several key issues still need to be solved. The first issue is forecasting a single time series, which does not capture the overall market trend and leaves the relationships between stocks and the mechanisms influencing price transmission unexplored. The second one is that constructing stock market graphs relies heavily on prior knowledge, and is challenging to adapt automatically. The third is relations obtained from the knowledge base are not effectively utilized and are only used as tools for building graph networks.

To address these limitations, we propose a framework called Dynamic Graph Representation with Contrastive Learning (DGRCL), which consists of three main components, namely embedding enhancement, contrastive constrained training, and dynamic graph evolution. In embedding enhancement, we adaptively construct dynamic edges that follow Zipf's law. Subsequently, the enhanced features are derived from OHLCV, computed with refined distance calculations utilizing the Fourier transform. Using these enhanced features and dynamic edges, we continue to apply GCNs to these evolving graphs to extract the latent node embeddings. To further refine the embeddings and optimize predictions, we integrate company relationships as constraints within an adaptive contrastive learning module. This module employs contrastive loss in the latent space to maximize the consistency between two augmented views of the input graph, thereby ensuring more robust feature representation. Finally, our model captures dynamic changes in the graph over time by employing a time series forecasting architecture, which can utilize either a Gated Recurrent Unit (GRU) or LSTM. This enables DGRCL to generate accurate and temporally-aware predictions.

To the best of our knowledge, our work is the first to incorporate stock relations as constraints into contrastive learning models for stock prediction, especially those based on GNNs. The main contributions of this work can be summarised as follows:
\begin{itemize}
    \item We adaptively construct dynamic edges that follow the objective laws, refining stock features through Fourier transform-based distance calculations. This allows the model to more effectively capture the intricate and dynamic relationships between stocks.
    \item We integrate company relationships as constraints within a contrastive learning module, employing contrastive loss in the latent space to maximize the consistency between two augmented views of the input graph. This ensures a robust feature representation that reflects the complex interdependencies among stocks.
    \item We conduct extensive experiments on real-world stock data from the US market.
    We conduct extensive experiments on real-world stock datasets with 2,763 stocks from two famous US markets (NASDAQ and NYSE). The experimental results demonstrate that DGRCL significantly outperforms state-of-the-art baselines in
    predicting the next trading day movement, with
    average improvements of 2.48\% in classification accuracy, 5.53 in F1 score, and 6.67
in Matthew correlation coefficient.
\end{itemize}

\section{\uppercase{Related Work}}
\label{sec: related work}

\subsection{GNNs}
\label{subsec:gnns}
GNNs are designed with specific consideration for tasks involving graph data, such as node classification, graph classification, and link prediction. In GALSTM \citep{yin2022graph}, an attention-based LSTM is constructed to learn weight matrices from Hawkes correlation graphs generated by different stocks to improve the stock selection winning rate while paying less attention to the optimal trading strategy. The most profitable stock prediction is also one task of portfolio management. Intra-sector relations, inter-sector relations, and stock-sector relations are considered in the previous study \citep{hsu2021fingat} to find hierarchical influence among stocks and predict the most profitable stock. Although sector-level modeling is performed in their model, some fine-grained metadata with descriptions of listed company attributes deserves more attention. Stock description documents can also be introduced into the model. For example, Yahoo Finance is one of the channels for topic-based extraction models in TRAN \citep{gao2021graph} to make stock recommendations. Similar limitation, TRAN only constructs the stock graph as a static one. Current approaches need to focus on portfolio construction and selection, as well as on understanding the dynamic interactions between these portfolios.

\subsection{Graph Contrastive Learning}
\label{subsec: graph contrastive learning}
Recently, contrastive representation learning on graphs has gained significant interest. It contrasts positive and negative sample pairs, aiming to bring similar samples closer and push dissimilar ones apart, focusing on solving the graph manual labeling problem in the real world. According to \citep{liu2022graph}, previous research can be categorized into cross-scale contrasting and same-scale contrasting. Cross-scale contrasting involves comparing elements at different scales. For example, MVGRL \citep{hassani2020contrastive} takes a multi-view approach by using a diffused graph as a global view and maximizing MI across both views and scales. DGI \citep{velivckovic2019deep} contrasts patch-level and graph-level representations by maximizing mutual information, which helps propagate global information to local representations. Same-scale contrasting focuses on elements at the same scale, such as comparing graphs directly. Within this approach, methods are further divided into context-based and augmentation-based. Context-based methods typically use random walks to identify positive pairs. Augmentation-based methods, like GraphCL \citep{you2020graph}, generate positive pairs through various perturbations at the graph level. GCA \citep{zhu2021graph} performs node-level augmentations tailored to the graph’s structure and attributes, while IGSD \citep{zhang2023iterative} uses graph diffusion to create augmented views and employs a teacher-student framework.

\section{\uppercase{Preliminary}}
\label{sec:preliminary}

\subsection{Notation}
\label{subsec: notation}
\subsubsection{Graph Representation Learning}
Let $\mathcal{G}=(\mathcal{V},\mathcal{E})$ denote the graph model where $\mathcal{V}$ is the set of nodes. The length of the time series is $T$, and $\bm{X}^{i}\in \mathbb{R}^{D\times T}$ is the covariate time-series data associated with node $i$, where each node is associated with $D$ different covariate time series. Furthermore, at time step $t$, covariate values for node $i$ is represented as $\mathbf{x}^{i}_{t}$, and $\bm{X}^{i}_{T} = \left \{ \mathbf{x}^{i}_{1} ,..., \mathbf{x}^{i}_{T} \right \}$ is the set of time-series in graph $\mathcal{G}$. Two different nodes $i$ and $j$ that contain an edge $\left(i,j\right) \in \mathcal{E} $ without a direction, forming graph $\mathcal{G}$ an undirected one, encoding an explicit dependence between nodes. Let $\bm{A} \in \mathbb{R}^{N\times N}$ denote this sparse adjacency matrix at a certain time. If $(i,j) \in{\mathcal{E}}$, then $\bm{A}_{ij}$ represents the weight of the edge (dependency) between node $i$ and $j$, and $\bm{A}_{ij}= 0$ when $(i,j) \notin \mathcal{E}$ otherwise. Hence, the graph structure is represented by the adjacency matrix $\bm{A}$ and its corresponding time-series $\left\{ \bm{X}^{i}_{T}| i=1, ...,N\right \}$.

\subsubsection{Graph Contrastive Learning}
Given an input Graph $\mathcal{G}$, graph contrastive learning aims to learn the representations in node-level tasks in this work by maximizing the feature consistency between two augmented views of the input graph via contrastive loss in the latent space. Data augmentation $q(\cdot )$ is a common operation to obtain two views  $v_{a}$ and $v_{b}$ for the same graph $\mathcal{G}$, including node dropping, edge perturbation, subgraph sampling, etc. With these views, latent representations $z_{a}$ and $z_{b}$ are then extracted by GNNs. Finally, given the latent representations, a contrastive loss is optimized to score the positive pairs $\left\{ z_{a}^{i}, z_{b}^{i} \right \}$ higher compared to other negative pairs, including inter-view negative pairs $\left\{z_{a}^{i}, z_{b}^{j}\right \}$, and intra-view negative pairs $\left\{z_{a}^{i}, z_{a}^{j}\right \}$.

\subsection{Problem Setting}
\label{subsec: problem setting}
The proposed framework solves the following graph-based time-series forecasting problem. For all stocks, a set of stock historical sequence data at day $t$ is represented by  $\mathcal{X} \in \mathbb{R}^{N\times T \times D}$, where $D$ is the dimension of features of one stock, the OHLCV. With a given lookback window of length $\delta$, the model can make predictions for the next day's ($t+1$) movements of the stocks, using the data of $\mathcal{X}_{t-\delta:t}$. The movement is calculated by $p^{t+1}_{c}-p^{t}_{c}$, in which $p^{t}_{c}$ means the \textit{close} price of the stock at day $t$, resulting two prediction outcomes. A positive result means the price will go up $(label = 1)$, while a negative result means it will go down $(label = 0)$, making this a binary classification task.

\section{\uppercase{Methodology}}
\label{sec:methodology}
\begin{figure*}[ht]
  \centering
  \includegraphics[width=1.0\linewidth]{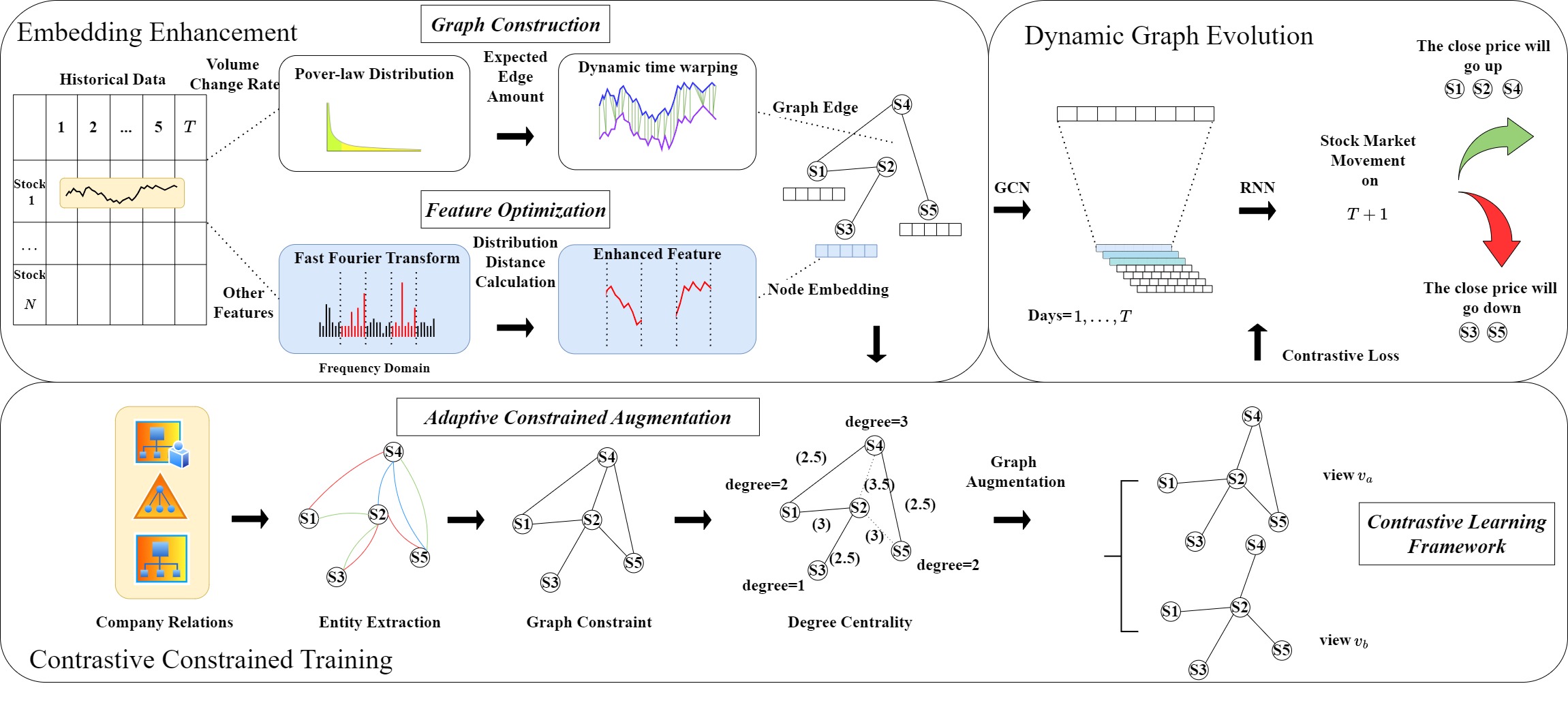}
  \caption{Schematic illustration of DGRCL framework. Top-left: In the process of embedding enhancement, dynamic stock graphs with enhanced features are generated. Bottom: Then, an encoder is trained with the company relations in a contrastive manner inside the CCT module, to improve the generation of an initial embedding matrix for the subsequent step. Top-right: Finally, the layer weights are learned through general RNNs (e.g., GRU, LSTM) to yield the final predictions.} 
  \label{fig:DGRCL}
\end{figure*}

In this section, we elucidate the framework of the proposed DGRCL as shown in Fig.\ref{fig:DGRCL}. Given OHLCV of the stocks in a certain interval $\left [ t-\delta+1, t\right ]$ and the company relations, we aim to predict the next day's stock price movement.  First, we construct the stock graph with envolving edges for different time steps and generate initial embeddings for each node. Second, we extract the dynamic node representations for each time step. Then, a constraint data augmentation method is encoded using the company relations. After that, we feed the constrained dynamic node representations to an RNN-based model to get the final representation of the nodes. Finally, we predict the stock movement based on the generated node representation from RNNs.

\subsection{Embedding Enhancement}
\label{subsec: embedding enhancement}

\subsubsection{Graph Construction}
The future performance of a stock is influenced by its past behavior and its connections to related stocks. Current research often depends heavily on established knowledge, such as manual industry classifications \citep{huynh2023efficient, feng2019temporal, xu2022hist}. However, this approach can be expensive, relies heavily on empirical experience, and may not generalize well to new forecasting tasks. In this regard, we follow the data-driven manner in \citep{wang2022adaptive} to calculate a proximity function by applying Dynamic Time Warping (DTW) \citep{jeong2011weighted} on the input signals $\mathbf{x}^{i}_{t}$, which will be further introduced in the experiment part. This process involves utilizing dynamic programming on a path matrix to determine the minimum alignment cost for each pair of stock sequences $\mathbf{x}^{i}_{t}$ and $\mathbf{x}^{j}_{t}$ with a lookback window $\delta$:

\begin{equation}
    cost(\mathbf{x}^{i}_{t} ,\mathbf{x}^{j}_{t} )\Longleftarrow DTW(\mathbf{x}^{i}_{t-\delta:t}, \mathbf{x}^{j}_{t-\delta:t})
\end{equation}

In DGRCL, we set the time step as 1. Through computations for each pair of stock sequences within the same time step, a corresponding cost matrix $\dot{\bm{A}}_{t}$ is generated for that particular time step $t$, and the total number of the cost matrices is $T-\delta$.

Traditional methods typically employ a hard threshold to partition the cost matrix $\dot{\bm{A}}_{t}$ and form the adjacency matrix $\bm{A}_{t}$ of the graph \citep{wang2022adaptive}. However, such approaches are often subjective. The adjacency matrix, which represents the edges in a graph, heavily relies on the structural information of its edges for information transmission. Hence, we introduce an adaptive method to estimate the number of significant edges that a graph should possess based on the number of nodes it contains.

According to \citep{sawhney2021exploring}, the scale-free structure of stock graphs is characterized by a power-law distribution in node degrees, at least in an asymptotic sense. Specifically, the fraction $P(n)$ of nodes in the graph having $n$ connections to other nodes follows the distribution \citep{barabasi1999emergence}, where $s$ is the exponent.

\begin{equation}
P(n)\sim n^{-s} 
\end{equation}

 This observation can be explained by Zipf's law \citep{newman2005power}, an empirical principle that tends to hold approximately when a list of measured values is sorted in descending order. The probability mass function of the Zipf distribution is a discrete probability distribution as,

\begin{equation}
 P(X=n)=\frac{{n}^{-s}}{ {\textstyle \sum_{i=1}^{N}(i^{-s}} )} 
\end{equation}

where $N$ is the total number of possible outcomes, $n$ is a specific outcome. Then we can give the expectation of a random variable $X$ with a Zipf distribution using the formula:

\begin{equation}
         E(X)= {\textstyle \sum_{n=1}^{N}}n\cdot  P(X=n)
\label{eq:ex}
\end{equation}

It should be noted that when applied, $n$ will represent the count of known nodes, signifying the number of stocks in the stock graph. This approach allows us to provide a rough estimate of the expected number of edges $E(n)$ in a stock network based on a given number of nodes $n$. Then, by employing a binary classification method, we manage the number of values exceeding the threshold to remain below the specified limit $E(n)$. This enables us to dynamically construct edges to build a stock graph $\mathcal{G}$. Furthermore, we obtain the dynamic graph for each time step as $\mathcal{G}_{1}, \mathcal{G}_{2}...\mathcal{G}_{T-\delta}$.

\subsubsection{Feature Optimization}
Statistical properties such as mean and variance in time series data often change over time, leading to distribution shift problems \citep{du2021adarnn, kim2021reversible}. This change in time distribution poses a significant challenge in stock series to achieving accurate time series forecasting. Therefore, we need to enhance the input stock series.

To identify the distribution characteristics of a time series, it needs to be transformed into another domain, such as the frequency domain. The Fourier transform (FT) \citep{cochran1967fast, bloomfield2004fourier} is an integral transform that presents a function into a representation that emphasizes the original function, which has wide applications in physics, mathematics, and engineering.

Here, we apply the Fast Fourier Transformation (FFT) \citep{cochran1967fast} to one stock series, and follow the temporal distribution characterization in \citep{du2021adarnn}, with the principle of maximum entropy \citep{jaynes1982rationale} to find the most dissimilar periods, whose objective can be formulated as:

\begin{equation}
    \max_{0<k\le k_{0}} \max_{n_{1},...,n_{k}} \frac{1}{k} \sum _{1<i\ne j\le k}d(FFT(\mathbf{x}^{i}), FFT(\mathbf{x}^{j}))
\end{equation}

\begin{equation}
    s.t. \forall i, \triangle_{1} < \left |  \mathbf{x}^{i} \right |  < \triangle_{2} ; \sum_{i}\left | \mathbf{x}^{i} \right | = \bm{X}^{i} \in \mathbb{R}^{D \times T}
\end{equation}

where $d(\cdot)$ is a distance metric, $k_{0}$ is the hyperparameter to avoid over-splitting, and $\triangle_{1}, \triangle_{2}$ are predefined parameters to avoid trivial solutions (e.g., extremely small or large values might not effectively capture the distribution information). $T$ is the length of time-series and $D$ is the dimension of features. The metric $d(\cdot)$ can be any distribution distance calculations such as Euclidean, JS-divergence, and KL-divergence. By searching $k$, the corresponding periods, and then applying this approach to each feature dimension $d$, we can find the most dissimilar period pairs in the feature domain for the stock. By performing similar operations on each stock, we obtain the enhanced stock graph node embedding matrix $\bm{X}'$.

It's important to notice that in one stock series, the sequences derived from FFT transformation in both rising and falling markets exhibit similarity, details please refer the experiments. The distinguishing factor lies in the fluctuating market. Therefore, we will generate three sets of results (rising, falling, and fluctuating) to serve as enhanced representations $\mathbf{x'}^{i}$ of one feature for stock $i$.

\subsection{Contrastive Constrained Training}
\label{subsec: contrastive constrained learning}
In previous stock prediction methods, company relations are mainly used to define the edge structure of stock graphs for information propagation \citep{feng2019temporal, kim2019hats}. Consideration of these relations as a constraint has been largely ignored during the training process. Therefore, we propose a Contrastive Constrained Training (CCT) module that integrates company relations to learn a constrained GNN encoder through contrastive learning. We will begin by introducing the adaptive constrained augmentation, followed by the contrastive learning framework, as the two components of the CCT module.

\subsubsection{Adaptive Constrained Augmentation}
Contrastive learning methods, which aim to maximize agreement between views, strive to learn representations consistent despite perturbations introduced by augmentation techniques. Within the CCT module, we propose an adaptive enhancement scheme based on company relation constraints. The aim is designed to disrupt less important links and features while protecting the important constraint structure and properties. 

In this step, contrary to previous methods, we incorporate company relations as constraints during training, resulting in a reevaluation of the former stock graph. The probability of the edge between node $i, j$ remains in the augmentation step $P\left \{ {(i,j)\in \mathcal{E}}  \right \}$ is calculated as: 

\begin{equation}
    1-min\left (\frac{\log{w}_{max}-\log{w}}{\log{w}_{max}-\log{w}_{\mu}} \mathrm{e} \cdot p_{e}, \quad p_{\tau }  \right ) 
\label{eq:pe and pt}
\end{equation}

where $w$ is the average of the degree centrality \citep{newman2018networks} between node $i$ and $j$. We first calculate $w$ for all edges and then make a max-mean normalization using the $max$ and mean ($\mu$). To balance the probabilities, we then take the logarithm of the result. $p_{e}$ and $p_{\tau}$ are hyperparameters to control the overall probability of removing edges and the cut-off threshold. 

The key approach presented in CCT is that the degree centrality calculation method is based on the company relation of the stocks, rather than the edges within the stock graph. That is, given the stock relation encodings $\bm{A}$ and each kind of relation $\bm{A}^{relation}_{ij} \in \mathcal{A}$ for node pair $(i, j)$ we must adhere to the constraint that there must be at least one type of relation that exists $\bm{A}^{\text{relation}}_{ij} > 0$. With this constraint, we can calculate $w$ using the following method.

\begin{equation}
    \frac{1}{2} \left(\sum_{(i,j) \in \mathcal{E}} \bm{A}^{\text{relation}}_{ji} + \sum_{(i,j) \in \mathcal{E}} \bm{A}^{\text{relation}}_{ij}\right)
\end{equation}

After calculating the edge remaining probability, we generate two views for the input graph $\mathcal{G}$ to perform contrastive learning. 

\subsubsection{Contrastive Learning Framework}
The proposed CCT module follows the common graph contrastive learning paradigm, in which the model aims to maximize the alignment of representations across various perspectives \citep{hassani2020contrastive, zhu2021graph}. We initially create two different views of the graph $\mathcal{G}$, by applying adaptive constrained augmentation to the input. Subsequently, we utilize a contrastive objective function that ensures the encoded embeddings of each node in these two distinct views align with each other, while also being distinguishable from embeddings of other nodes.

Specifically, during each training epoch, the CCT module begins by randomly selecting two data augmentation functions from $q(\cdot )$, which is a common operation to obtain two views  $v_{a}\sim q(G)$, $v_{b}\sim q(G)$, then, node representations $z_{a}$ and $z_{b}$ are obtained from a GNN encoder $g(\cdot)$. For a given node $i$, its embedding $z^{i}_{a}$ generated in view $v_{a}$ serves as the anchor, while the embedding of it generated in the other view $z^{i}_{b}$ forms the positive sample, and the remaining embeddings in the two views are considered as negative samples. Following \citep{zhu2021graph}, we define the pairwise objective $\ell (z^{i}_{a}, z^{j}_{b})$ for each positive pair $(z^{i}_{a}, z^{i}_{b})$ as :

\begin{equation}
\log \frac{e^{\theta (z^{i}_{a}, z^{i}_{b})/\tau }}{e^{\theta (z^{i}_{a}, z^{i}_{b})/\tau }+\sum{e^{\theta (z^{i}_{a}, z^{j}_{b})/\tau }}+ \sum{e^{\theta (z^{i}_{a}, z^{j}_{a})/\tau }}} (i\ne j)
\end{equation}

where $\tau$ is a temperature parameter, and $e^{\theta (z^{i}_{a}, z^{i}_{b})/\tau }$ represent positive pair. $\sum{e^{\theta (z^{i}_{a}, z^{j}_{b})/\tau }}, \sum{e^{\theta (z^{i}_{a}, z^{j}_{a})/\tau }}$ mean inter-view negative pairs and intra-view negative pairs. $\theta(z_{a}, z_{b})= s(\phi(z_{a}), \phi(z_{b}))$, in which, $s(\cdot)$ refers to the cosine similarity and $\phi(\cdot)$ is a non-linear projection, implemented as a two-layer perception model. Moreover, the overall objective $\mathcal{L}_{cl}$ for the CCT module to be maximized uses the average for all positive pairs, since the two views are symmetric. 

\begin{equation}
    \mathcal{L}_{cl}=\frac{1}{2N}\sum_{i=1}^{N}{(\ell (z^{i}_{a}, z^{i}_{b})+\ell (z^{i}_{b}, z^{i}_{a}))} 
\end{equation}

In this way, we use the number of relationships as a basis for measuring the possibility of the existence of a certain edge, thereby achieving adaptive data augmentation with constraints, which is then trained together with the dynamic graph evolution.

\subsection{Dynamic Graph Evolution}
\label{subsec: dynamic graph evolution}
Using dynamic graphs and node embedding matrices, we will now explain how to acquire the dynamic latent representations of nodes. This process consists of two components: the graph convolutional network and the weight evolution.

Based on GCN \citep{kipf2017semisupervised}, we propose the following propagation rules:

\begin{align}
    \bm{H}^{(l+1)}_{t}=\sigma 
    (\widehat{\bm{A}}_{t}  \bm{H}^{(l)}_{t}\bm{W}^{(l)}_{t})
\end{align}

Here, at time $t$, the $l$-th layer takes the adjacency matrix $\bm{A}_{t}$ and the node embedding matrix $\bm{H}^{(l)}_t$ as input, and outputs $\bm{H}^{(l+1)}_{t}$ using a layer-specific trainable weight matrix $\bm{W}^{(l)}_t$ to update. Specifically, $\sigma(\cdot)$ denotes an activation function.

\begin{equation}
    \widehat{\bm{A}} = \widetilde{\bm{D}}^{-\frac{1}{2}}\widetilde{\bm{A}}\widetilde{\bm{D}}^{-\frac{1}{2}},
    \widetilde{\bm{A}}=\bm{A}+\bm{I},
    \widetilde{\bm{D}}=diag\left ( \sum_{j}\widetilde{\bm{A}}_{ij}\right ) 
\end{equation}

To acquire the dynamic latent representations, the trainable weight matrix $\bm{W}^{(l)}_t$, we follow the methodology outlined in \citep{pareja2020evolvegcn}. $\bm{W}^{(l)}_{t}$ is treated as the hidden state of the dynamical system updated by any RNN-based model $f(\cdot)$, such as LSTM and GRU, with a prediction loss $\mathcal{L}_{pred}$.

\begin{equation}
    \bm{W}^{(l)}_{t}=f(\bm{H}^{(l)}_{t}, \bm{W}^{(l)}_{t-1})
\end{equation}

The final prediction result will be derived from the output of $\bm{W}$ at the final time step. This result is binarized and then mapped through a $softmax$ function to predict each stock. The final form of the loss function is shown below, where $\lambda$ is a trade-off parameter.

\begin{equation}
    \mathcal{L}=\mathcal{L}_{pred}+\lambda\mathcal{L}_{cl}
\end{equation}

\section{\uppercase{Experiment}}
\label{sec:experiment}
In this section, we describe the experimental setup used to evaluate the effectiveness of our proposed method. Experimental results demonstrate the effectiveness of our DGRCL framework in comparison with different backbones and datasets. We specifically aim to answer the following questions: \textbf{(RQ1)}: How effective is the proposed DGRCL framework for the graph classification task? \textbf{(RQ2)}: How effective are the individual modules proposed in DGRCL? \textbf{(RQ3)}: How does sensitivity to hyperparameters affect the performance of the proposed model?

\subsection{Dataset}
\label{subsec: dataset}
We closely follow the experimental setup outlined in \cite{feng2019temporal}, using two major US stock markets. Many studies have utilized this same dataset to evaluate their stock forecasting models \citep{kim2019hats, sawhney2021stock, hsu2021fingat, you2024dgdnn, you2024multi}, and to ensure easier comparison with the existing literature, we also adopt this dataset in our experiments.

Dataset statistics are summarized in Table \ref{table:dataset}. We follow the setup of \citep{pareja2020evolvegcn} and divide the two datasets into training, validation, and testing sets in the proportions of 0.65, 0.1, and 0.25, respectively. This results in 611, 94, and 236 effective trading days for each set. OHLCV are daily values for each stock. Edges number is calculated by \eqref{eq:ex} with a certain number of nodes. Recognizing that stocks within the same industry tend to be similarly affected by industry-wide prospects, we gather data on the sector-industry relationships among stocks. For two datasets, we apply the sector-industry relations as Relation Types in Table \ref{table:dataset}, for more details, please refer to \citep{feng2019temporal}. The dataset and source code are available in github.\footnote{https://anonymous.4open.science/r/DGRCL-359C/README.md}

Each stock market dataset spans four years, exceeding one thousand time points, with more than one thousand traded stocks in each market. Compared to the datasets used in the baseline models, the datasets employed in DGRCL are characterized by their long duration, multiple nodes, and dynamic temporal graph changes. This makes financial datasets particularly well-suited for studying long-term information propagation in GNNs, as compared to other datasets.

\begin{table}[tb]
\centering
\caption{Data statistics of NASDAQ and NYSE.}
\resizebox{0.5 \textwidth}{!}{%
\begin{tabular}{@{}lcccccc@{}}
\toprule
\multirow{2}{*}{Item}   & \multicolumn{3}{c}{NASDAQ dataset}        & \multicolumn{3}{c}{NYSE dataset}          \\ \cmidrule(l){2-7} 
                        & Train         & Valid        & Test       & Train        & Valid        & Test        \\ \midrule
Time Span               & \multicolumn{3}{c}{01/02/2013-12/08/2017} & \multicolumn{3}{c}{01/02/2013-12/08/2017} \\
Valid Trading Days      & 611           & 94           & 236        & 611          & 94           & 236         \\
Data Points             & \multicolumn{3}{c}{5 (OHLCV)}              & \multicolumn{3}{c}{5 (OHLCV)}              \\
Nodes                   & \multicolumn{3}{c}{1026}                  & \multicolumn{3}{c}{1737}                  \\
Edges                   & \multicolumn{3}{c}{164}                  & \multicolumn{3}{c}{255}                   \\
Relation Types          & \multicolumn{3}{c}{112}                   & \multicolumn{3}{c}{130}                   \\ \bottomrule
\end{tabular}%
}
\label{table:dataset}
\end{table}

\subsection{Model Setting}
\label{subsec: model setting}
All experiments are performed on a Nvidia GeForce RTX 3090 graphic card, CUDA version 11.5. The lookback window $\delta$ applied in $DTW(\cdot)$ is 20. In general, each month consists of approximately 22.5 working days. The input signals $\mathbf{x}^{i}_{t}$ for DTW is the volume volatile, by calculating $vol^{t}/ vol^{t-1}$. The RNN-based model $f(\cdot)$ is a two-layer LSTM, following the methodology outlined in \citet{pareja2020evolvegcn}. Three hyperparameters in the CCT module are $\tau$, $p_{e}$, and $p_{\tau}$. Following the work of \citet{zhu2021graph}, we set $\tau$ as 0.4. For the other two, we train the model using $p_{e}$ and $p_{\tau}$ from 0.1 to 0.9 with increments of 0.2. The activation function $\sigma(\cdot)$ is set to $ReLU(\cdot)$. Based on the method outlined in \citet{liu2022contrastive}, we control the trade-off parameter $\lambda$ to bring $\mathcal{L}_{pred}$ and $\mathcal{L}_{cl}$ to the same order of magnitude. The empirical value of $\lambda$ obtained from multiple experiments is 0.1.

\subsection{Baseline}
\label{subsec: baseline}
We consider state-of-the-art dynamic graph node classification models as baseline methods to learn the representations of nodes in a graph, including Graph WaveNet \citep{wu2019graph}, 
MTGODE \citep{jin2022multivariate},
STGCL \citep{liu2022contrastive}, and
Evolvegcn \citep{pareja2020evolvegcn}.

\begin{table*}[tb]
\centering
\caption{Performance of proposed DGRCL and other baselines on next trading day stock trend classification over the test period. Bold denotes the best result.}
\resizebox{\textwidth}{!}{%
\begin{tabular}{@{}lcccccc@{}}
\toprule
\multirow{2}{*}{Methods} & \multicolumn{3}{c}{NASDAQ dataset}                                                               & \multicolumn{3}{c}{NYSE dataset}                                                                 \\ \cmidrule(l){2-7} 
                              & Accuracy(\%)       & F1 score            & MCC(\(\times 10^{-3}\))       & Accuracy(\%)       & F1 score            & MCC(\(\times 10^{-3}\)) \\ \midrule
Graph WaveNet \citep{wu2019graph}
                 & 50.94±0.15          & 60.51±1.10          & 5.50±1.00                            & 50.86±0.13          & 57.28±1.86          & 11.12±2.21                     \\
MTGODE \citep{jin2022multivariate}                       & 51.41±0.15          & 62.41±4.24          & 5.92±2.61                        & 52.35±0.17          & 62.73±1.72          & 20.62±5.76                     \\
STGCL \citep{liu2022contrastive}                        & 50.56±0.08          & 57.81±1.19          & \textbf{9.71±1.15}                  & 51.21±0.05          & 67.47±0.07          & 22.33±1.53                     \\
EvolveGCN \citep{pareja2020evolvegcn}                 & 50.19±0.02          & 56.94±0.21          & 4.41±0.76                         & 51.14±0.15          & 66.85±0.37          & 14.41±0.45                     \\ \midrule
DGRCL (Ours)              & \textbf{53.06±0.18} & \textbf{66.53±2.74} & 9.32±1.53                      & \textbf{54.07±0.20} & \textbf{67.53±0.19} & \textbf{27.51±5.52}            \\ \bottomrule
\end{tabular}
}
\label{table:performance}
\end{table*}

\begin{table*}[htb]
\centering
\caption{Ablation study on the two datasets. We replace the Embedding Enhancement module and Contrastive Constrained Training module in DGRCL with their discrete implementations, denoted w/o EE and w/o CCT. We further remove all proposed modules except dynamic graphs for training as w/o EE \& CCT.}
\resizebox{\textwidth}{!}{%
\begin{tabular}{@{}lcccccc@{}}
\toprule
\multirow{2}{*}{Methods} & \multicolumn{3}{c}{NASDAQ dataset}                          & \multicolumn{3}{c}{NYSE dataset}                            \\ \cmidrule(l){2-7} 
                         & Accuracy(\%) & F1 score   & MCC(\(\times 10^{-3}\)) & Accuracy(\%) & F1 score   & MCC(\(\times 10^{-3}\)) \\ \midrule
DGRCL                    & 53.06±0.18    & 66.53±2.74 & 9.32±1.53                      & 54.07±0.20    & 67.53±0.19 & 27.51±5.52                     \\ \midrule
w/o EE                   & 51.45±0.03    & 62.48±1.19 & 4.69±0.02                      & 52.77±0.05    & 62.95±1.40 & 21.88±2.35                     \\
w/o CCT                  & 52.69±0.32    & 65.61±0.22 & 5.78±1.07                      & 53.87±0.11    & 67.34±0.18 & 23.12±1.77                     \\
w/o EE \& CCT            & 51.08±0.12    & 57.43±1.22 & 2.91±0.26                      & 51.65±0.21    & 64.12±0.19 & 7.61±2.11                      \\ \bottomrule
\end{tabular}
}
\label{table:ablation}
\end{table*}

\begin{itemize}
    \item Graph WaveNet \citep{wu2019graph}:
    A spatiotemporal graph modeling approach that captures spatial-temporal dependencies across multiple time series by integrating graph convolution with dilated causal convolution. Two datasets are used, the METR-LA dataset spans four months of traffic speed data from 207 sensors, while the PEMS-BAY dataset covers six months from 325 sensors, both aggregated into 5-minute intervals.
    \item MTGODE \citep{jin2022multivariate}:
    A continuous model that forecasts multivariate time series by integrating dynamic graph neural Ordinary Differential Equations to unify spatial-temporal message passing and enhance latent spatial-temporal dynamics. The evaluation involves five benchmark datasets: three conventional time series datasets (Electricity, Solar-Energy, Traffic) and two traffic datasets (Metr-La, Pems-Bay) with predefined graph structures and specific time spans. 
    \item STGCL \citep{liu2022contrastive}:
    A study that integrates contrastive learning into spatio-temporal graph (STG) forecasting, leveraging novel node and graph-level contrastive tasks to mitigate data scarcity, enhancing performance through joint learning schemes and strategic data augmentations. The model is tested using two traffic benchmarks, PEMS-04 and PEMS-08, with aggregated 5-minute traffic flow, average speed, and occupancy data, applying Z-score normalization and Gaussian kernel adjacency matrices.
    \item EvolveGCN \citep{pareja2020evolvegcn}:
    An approach to dynamic graph representation learning that adapts graph convolutional networks over time by evolving GCN parameters with recurrent neural networks, without relying on node embeddings. The benchmark datasets include the Stochastic Block Model (SBM) for simulating community structures, Bitcoin OTC (BC-OTC) and Bitcoin Alpha (BC-Alpha) for rating predictions, UC Irvine messages (UCI) for link prediction, and several other networks for various predictive tasks.
\end{itemize}

\subsection{Evaluation Metric}\label{subsec: evaluation metric}
Following the approaches taken in previous studies \citep{you2024dgdnn, kim2019hats, deng2019knowledge, sawhney2020deep}, we assess our result using Classification Accuracy, F1 score, and Matthews Correlation Coefficient (MCC) to evaluate model performance. For all three metrics, higher values indicate better model performance. We record the time taken for each training epoch of the models in seconds.

\subsection{Evaluation Result}\label{subsec: evaluation result}

\subsubsection{RQ1. Performance Comparison}
  
The experimental results in Table \ref{table:performance} show that our model outperforms all other baseline models in terms of accuracy and F1 score, and is also comparable to the baseline models in terms of MCC. For the NASDAQ dataset, DGRCL achieves the highest accuracy at 53.06\%, surpassing the second-best MTGODE by 1.65\%. Additionally, DGRCL attains the top F1 score of 66.53\% and a strong MCC value of 9.32, nearly matching the best MCC of 9.71. On the NYSE dataset, DGRCL similarly outperforms other methods, with an accuracy of 54.07\%, an F1 score of 67.53\%, and a substantial MCC of 27.51, marking the highest recorded metrics across all models. The notable discrepancy where MTGODE exhibits a high F1 score but a lower MCC indicates the model may handle accuracy and recall well for the majority class but struggles with the minority class, affecting its overall prediction quality. A similar finding is also shown in the performance of EvolveGCN on the NYSE dataset.

Moreover, we observe that the accuracy of all baseline methods is not very high, being only slightly better than 50\%, which deviates significantly from the experimental results described in their respective papers. A reasonable explanation for this inconsistency is that the nodes and the temporal span of their datasets are relatively small. As a result, their models may be unable to effectively learn the representations of large-scale graph models over long time series. For instance, in the datasets used by Graph WaveNet and STGCL, the number of nodes is 207 and 325, respectively, which is far smaller than the 1026 and 1737 nodes in our dataset. Similarly, in the dataset used by MTGODE, the largest sample size is 52,560, which is much smaller than the smallest sample size in our study 626,886 (1026$\times$611).


\begin{figure*}[tb]
  \centering
    \includegraphics[width=0.45\linewidth]{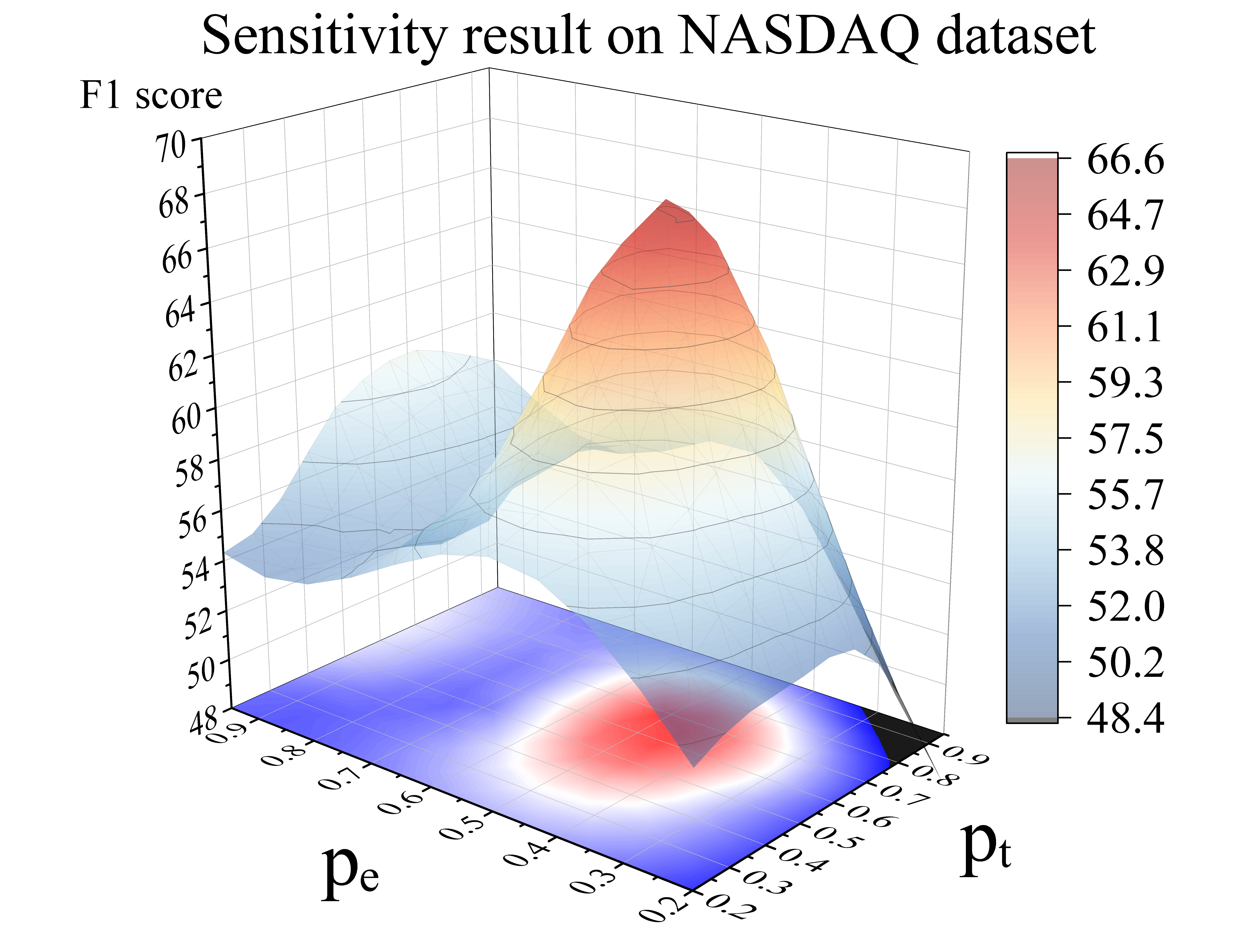}
    \includegraphics[width=0.45\linewidth]{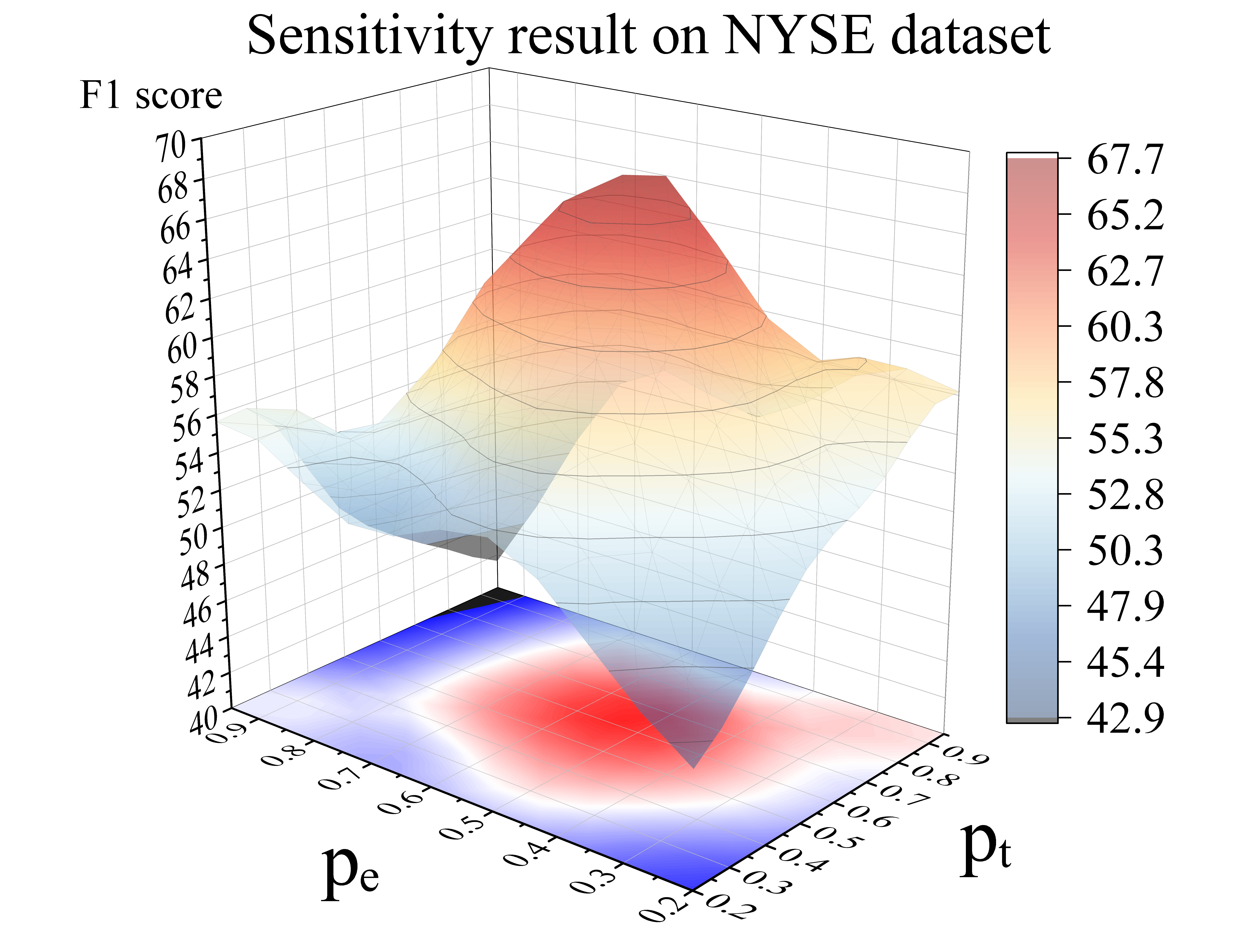}

  \caption{Sensitivity study of the overall probability of removing edges $p_{e}$ and the cut-off threshold $p_{\tau}$ via F1 score.}
  \label{fig:sensitivity}
\end{figure*}


  
  
  

\begin{table*}[tb]
\centering
\caption{Run-time complexity of the experiments}
\resizebox{\textwidth}{!}{%
\begin{tabular}{lcc|lcc}
\hline
\multicolumn{1}{c}{Model} & Dataset & Epoch Training Time (s) & \multicolumn{1}{c}{Model} & Dataset & Epoch Training Time (s) \\ \hline
Graph WaveNet             & NASDAQ  & 1.63                   & DGRCL (Ours) w/o EE        & NASDAQ  & 406.53                 \\
                          & NYSE    & 3.14                   &                           & NYSE    & 485.38                 \\
MTGODE                    & NASDAQ  & 13.22                  & DGRCL (Ours) w/o CCT       & NASDAQ  & 42.58                  \\
                          & NYSE    & 29.96                  &                           & NYSE    & 96.43                  \\
STGCL                     & NASDAQ  & 2.12                   & DGRCL (Ours) w/o EE \& CCT & NASDAQ  & 41.79                  \\
                          & NYSE    & 3.17                   &                           & NYSE    & 80.06                  \\
EvolveGCN                 & NASDAQ  & 24.63                  & DGRCL (Ours)               & NASDAQ  & 414.05                 \\
                          & NYSE    & 38.25                  &                           & NYSE    & 505.54                 \\ \hline
\end{tabular}
}
\label{table:complexity}
\end{table*}

\subsubsection{RQ2. Ablation Study}

To verify the effectiveness of each component of DGRCL, we conducted ablation studies on the two datasets (see Table~\ref{table:ablation}). The complete DGRCL model incorporates EE and CCT, exhibiting the highest performance in all metrics. When both EE and CCT are removed (w/o EE \& CCT), DGRCL has the most significant performance drop, with accuracy falling to 51.08\% and 51.65\%, the F1 score to 57.43 and 64.12, and the MCC to 2.91 and 7.61 in the NASDAQ and NYSE datasets respectively. 

From the result we can find that, by incorporating the EE module, DGRCL experiences a significant improvement, nearly doubling the MCC score on both datasets. This indicates that the EE module enables the model to learn more effective latent features of the time series, thereby enhancing its predictive performance. Another observation is that the CCT module alone does not result in a significant improvement in model performance and is less effective compared to adding the EE module alone. Nonetheless, when both modules are combined, they greatly enhance the overall performance of the model. This suggests that the model achieves optimal predictive results by learning the dynamic temporal relationships and the static relationships in the graph data.

\subsubsection{RQ3. Hyperparameter Sensitivity}

We explore the sensitivity of two important hyperparameters: the overall probability of removing edges $p_{e}$ and the cutoff threshold $p_{\tau}$. 
These two hyperparameters directly affect the performance of the CCT module and further impact the overall performance of the DGRCL. Here, we use the global evaluation metric F1 score to reflect the overall performance of different combinations of these hyperparameters. Results are presented in Fig.~\ref{fig:sensitivity}. We make the following observations.

When both $p_{e}$ and $p_{\tau}$ are set at their maximum or minimum values simultaneously, overall performance is poor, with the F1 score in both datasets falling below 50. This indicates that finding a balance between these two hyperparameters is crucial to improving the model's predictive performance. The second finding is that $p_{e}$ is more important than $p_{\tau}$, as it determines the dynamic lower bound in Eq.~\eqref{eq:pe and pt}. If $p_{\tau}$ is set very low, $P\left \{ {(i,j)\in \mathcal{E}}  \right \}$ becomes a fixed value equal to $1-p_{\tau}$. This indicates that all edges have the same existence probability, resulting in random remaining. Another observation is that, Compared to the NASDAQ dataset, the NYSE dataset has a larger optimal hyperparameter space, making it easier to find local optima during model training. Previous financial papers \citep{lian2022empirical, jiang2011comparison} also support this finding, noting that the NASDAQ market is more volatile, leading to a smaller optimal hyperparameter space.

\subsubsection{Complexity}

Table~\ref{table:complexity} compares the training times per epoch (in seconds) of various models on the NASDAQ and NYSE datasets. Our model, DGRCL, generally has significantly higher training times compared to other models. For example, Graph WaveNet and STGCL require only 1.63s and 2.12s per epoch on NASDAQ, respectively, while DGRCL without embedding enhancement (EE) and contrastive constrained training (CCT) takes 41.79s. The full DGRCL model, which includes these enhancements, requires 414.05s on NASDAQ and 505.54s on NYSE.

Notably, adding the CCT unit significantly increases the training time, accounting for around 90\% of the total run-time, which indicates that the graph contrastive learning component is the main source of computational overhead. However, as shown in the comparison between Table~\ref{table:performance} and Table~\ref{table:ablation}, even without the CCT module, DGRCL still outperforms the baselines in terms of accuracy. This demonstrates that our framework performs well without relying on the computationally expensive CCT module. While DGRCL offers improved prediction performance, this high computation cost, particularly from the CCT module, suggests a need for further optimization to balance performance and efficiency, especially in large-scale applications.

\section{\uppercase{Conclusion}}
\label{sec:conclusion}
In this paper, we propose the Dynamic Graph Representation with Contrastive Learning (DGRCL) framework, which unifies dynamic temporal graph learning and static relational graph learning through a contrastive learning paradigm. Within DGRCL, both the Embedding Enhancement (EE) and Contrastive Constrained Training (CCT) modules complement each other: the EE module dynamically captures evolving stock market trends, while the CCT module leverages static inter-stock relationships to refine feature representations through contrastive learning. This dual approach enhances the model’s ability to predict stock movements with higher accuracy. 
Additionally, we utilize objective laws to construct dynamic graph structures. Meanwhile, we integrate a Fourier transform-based technique into the EE module to effectively address the challenges posed by distribution shifts in time series data. Ultimately, we validate the effectiveness of DGRCL through extensive experiments on two real-world datasets, demonstrating that DGRCL consistently outperforms four baseline models.

For future work, we will also refine the CCT module to better distinguish between static and dynamic graph features. We will perform comparisons with additional baselines on financial market datasets from various countries (such as the Shanghai Stock Exchange in China) to further demonstrate the superiority of the DGRCL framework as a long-term time-series forecasting model for financial predictions. In addition, to make our framework more efficient, future work will focus on optimizing the graph contrastive learning part to reduce its computational overhead and enhance the overall performance of the DGRCL framework.

\ifnum\BLIND=0
{
    \section*{\uppercase{Acknowledgements}}

    This work was supported by UKRI EPSRC Grant No. EP/Y028392/1: AI for Collective Intelligence (AI4CI), and Innovate UK Project No. 10094067: Stratlib.AI - A Trusted Machine Learning Platform for Asset and Credit managers. The authors have no conflicts of interest to declare. 
}
\fi

\balance
\bibliographystyle{apalike}
{\small
\bibliography{icaart-25}}



\end{document}